\pdfoutput=1
\documentclass[11pt]{article}

\usepackage[final]{acl}
\usepackage{epstopdf}

\usepackage{times}
\usepackage{latexsym}
\usepackage{booktabs}
\usepackage{float}
\usepackage{makecell} 
\usepackage{array}  
\usepackage{amsfonts}

\usepackage{tablefootnote}
\usepackage{hyperref} 

\usepackage{svg}
\usepackage{caption}
\usepackage{subcaption}
\usepackage{afterpage}

\usepackage{booktabs}   % for \toprule, \midrule, \bottomrule
\usepackage{amssymb}    % for \checkmark
\usepackage{pifont}     % for \ding and \xmark
\newcommand{\cmark}{\ding{51}} % bold check
\newcommand{\xmark}{\ding{55}}
\usepackage{bm}

\usepackage{multirow}
\usepackage{multicol}
\usepackage{threeparttable} 

\usepackage[T1]{fontenc}
 \usepackage{amsmath}
\usepackage[utf8]{inputenc}

\usepackage{microtype}

\usepackage{inconsolata}

\usepackage{graphicx}
\usepackage{soul}

\usepackage{tikz}
\usetikzlibrary{shapes.geometric, arrows.meta, positioning, fit, backgrounds}
\usepackage{enumitem}
\usepackage[most]{tcolorbox}
\usepackage{xcolor} % For color definitions
\title{GraDeT-HTR: A Resource-Efficient Bengali Handwritten Text Recognition System utilizing Grapheme-based Tokenizer and Decoder-only Transformer}

\author{
 \textbf{Md.\ Mahmudul Hasan}\thanks{Equal contribution. Author order is randomly determined.},
 \textbf{Ahmed Nesar Tahsin Choudhury}\footnotemark[1],
\\
 \textbf{Mahmudul Hasan},
 \textbf{Md.\ Mosaddek Khan},
\\
 Computer Science and Engineering, University of Dhaka
 \\
  \texttt{\{mdmahmudul-2020215620, ahmednesartahsin-2020115612,}\\
\texttt{mahmudul-2019917803\}@cs.du.ac.bd, mosaddek@du.ac.bd}
}

\begin{document}
\maketitle

% WORKFLOW
% 1. Previous year EMNLP paper of the same track
% 2. Track Story of the Paper
% 3. Content Comparable to that paper

\begin{abstract}

Despite Bengali being the sixth most spoken language in the world, handwritten text recognition (HTR) systems for Bengali remain severely underdeveloped. The complexity of Bengali script—featuring conjuncts, diacritics, and highly variable handwriting styles—combined with a scarcity of annotated datasets makes this task particularly challenging. We present \textbf{GraDeT-HTR}\footnote{\url{https://cognistorm.ai/hcr}}, a resource-efficient Bengali handwritten text recognition system based on a \textbf{Gra}pheme-aware \textbf{De}coder-only \textbf{T}ransformer architecture. To address the unique challenges of Bengali script, we augment the performance of a decoder-only transformer by integrating a grapheme-based tokenizer and demonstrate\footnote{\url{https://github.com/mahmudulyeamim/GraDeT-HTR}} that it significantly improves recognition accuracy compared to conventional subword tokenizers. Our model is pretrained on large-scale synthetic data and fine-tuned on real human-annotated samples, achieving state-of-the-art performance on multiple benchmark datasets.

%Although Bengali is the sixth most spoken language in the world, the absence of comprehensive Bengali Handwritten Text Recognizers hampers the digitization of handwritten documents. Handwritten text recognition is a longstanding research challenge due to the inherent complexity of the task and the scarcity of annotated real-world data. Existing text recognition methods include CNN-RNN-based architectures and, more recently, transformer-based models. In this paper, we present \textbf{BnDTrOCR}, a comprehensive handwritten text recognition system for Bengali that leverages decoder-only transformers for reliable recognition performance. Due to the scarcity of annotated real-world data, the model is pre-trained with extensive synthetic data and fine-tuned with human-labeled real data. Additionally, we investigate the impact of grapheme-based tokenizers on text recognition in complex scripts like Bengali. Our experiments demonstrate that selecting an optimal tokenizer can both enhance performance and reduce the number of model parameters significantly. The system is available as a web-based interface\footnote{\url{https://cognistorm.ai/hcr}} and the model source code is available on GitHub\footnote{\url{https://github.com/username/repo}} under the MIT license.
\end{abstract}

\section{Introduction}
\label{sec:introduction}

Optical Character Recognition (OCR) involves transforming printed or handwritten images into machine-encoded text, supporting tasks such as digitizing pages from books, scene photographs, receipts, or notes. Within OCR, Handwritten Text Recognition (HTR) is focused specifically on transcribing handwritten content. HTR remains a challenging task due to the vast diversity in individual writing styles, inconsistent lighting, varying stroke width, cursive scripts, and artifacts such as smudges or noise.

Despite significant progress in handwritten text recognition (HTR) for languages such as English and Chinese, Bengali remains notably underserved. This gap is especially critical given the widespread use of handwritten Bengali in education, administration, and historical records. Developing effective HTR systems for Bengali presents unique challenges: the script contains visually complex ligatures, numerous diacritics, and allographic variations, all of which are compounded by highly diverse individual handwriting styles. These linguistic and visual complexities, combined with a scarcity of large, high-quality annotated datasets, make Bengali HTR a particularly demanding low-resource OCR task.

% \begin{figure}
%     \centering
%     \input{pipeline}
%     \caption{Overview of our OCR pipeline, showing the flow from input image through text detection and segmentation to final output.}
%     \label{fig:pipeline}
% \end{figure}

% [Citations to be added] Existing approaches to solve text recognition involve encoder-decoder architectures. The encoder extracts visual information from images, and the decoder produces the transcripted text. Early notable approaches involve CNNs as encoders, and RNNs with Connectionist Temporal Classification (CTC) as decoders. With the advent of transformers [cite Vaswani et al], researchers have tried to incorporate attention in both the encoder and decoder layers. Initial successful approaches involve replacing the RNN-based decoder with transformer-based language models [cite ...]. However, these approaches still used CNN-based architectures as encoders. [cite] TrOCR first tried to use pre-trained Vision Transformers [cite] in the encoder layer to use the pre-trained knowledge. More recently, [cite] DTrOCR tried to eliminate the encoder layer altogether and use only a decoder-only transformers approach to achieve SOTA performance. Based on the immense success of decoder-only transformers in OCR tasks, we use this architecture for Bengali text recognition. 

Existing approaches to text recognition predominantly employ encoder-decoder architectures, wherein the encoder is responsible for extracting visual features from images while the decoder generates the corresponding transcriptions. Early methods \citep{shi2016end2end} commonly utilized CNNs as encoders and RNNs, with CTC~\citep{graves2006}, as decoders. With the introduction of transformers \citep{vaswani2017attention}, attention-based mechanisms have been utilized to gain high performance in OCR tasks \citep{wang2019scene, sheng2019nrtr, zhang2020sahan, li2023trocr, fujitake2024dtrocr}.

% The introduction of transformers \citep{vaswani2018} has led researchers to incorporate attention mechanisms into both the encoder and decoder components \citep{zhang2020sahan, wang2019scene, sheng2019nrtr}. TrOCR \citep{li2023trocr} was the first to leverage pre-trained Vision Transformers \citep{dosovitskiy2021image, touvron2021training} within the encoder, thereby utilizing extensive pre-trained visual knowledge. More recently, DTrOCR \citep{fujitake2024dtrocr} proposed a decoder-only transformer architecture, eliminating the encoder altogether and achieving state-of-the-art performance.

In the context of Bengali, existing research on text recognition is relatively limited. Recent studies have explored architectures based on Convolutional Neural Networks (CNNs) and Recurrent Neural Networks (RNNs). However, the majority of these works \citep{azad2020bangla, hossain2021bangla, safir2021end, chaudhury2022deep} evaluate their methods on specific datasets, resulting in limited generalizability of their results. Some recent efforts \citep{hossain2022lilaboti} have considered evaluation on unseen datasets, thereby addressing issues of generalization to a certain extent. Nonetheless, approaches based on transformer architectures have not yet been explored for HTR tasks. For text detection, models such as BN-DRISHTI \citep{jubaer2023bn} attain satisfactory results, so the focus has largely shifted to the recognition stage. In parallel, recent work on tokenizer design \citep{basher2023bngraphemizer} has demonstrated that grapheme-based tokenization can improve recognition performance by addressing Bengali's intricate ligatures, conjuncts, numerous diacritics, and allographic variations.

Motivated by the strong performance of decoder-only transformer models in English and Chinese OCR tasks \citep{fujitake2024dtrocr}, as well as the potential significance of tokenizer selection, we propose a Bengali Handwritten Text Recognition system that leverages a decoder-only transformer architecture in conjunction with a grapheme-based tokenizer. The core contributions of this paper are summarized below:

% The core contributions of this paper are summarized below:
% \begin{itemize}
%     \item We present a state-of-the-art pipeline for Bengali Handwritten Text Recognition (HTR), which, to our knowledge, is the first to provide an \textit{open-source} end-to-end general solution integrating both text detection and recognition for full-page images.
%     \item To the best of our knowledge, we are the first to replace the default tokenizer of a transformer-based LLM with a handwriting-specific tokenizer and demonstrate superior performance in HTR tasks.
%     \item We show that strong performance can be achieved without large-scale pre-training of transformer-based LLMs, indicating the potential for effective HTR in low-resource language settings.
% \end{itemize}

\begin{itemize}[noitemsep,topsep=0pt]
  \item We develop an end-to-end Bengali HTR pipeline that integrates both text detection and recognition for full-page images.  
  \item  We study the effects of replacing the default tokenizer in a transformer-based model with a grapheme-based tokenizer tailored for Bengali—a direction that, to the best of our knowledge, has not been previously explored—and empirically show improvements in recognition performance.
  
  \item  We demonstrate that strong HTR performance can be achieved without relying on large-scale pretrained language models, offering a practical path for developing this demanding application in low-resource settings.
\end{itemize}

\section{Related Work}

This section reviews prior research in three key areas relevant to our work: general optical character recognition (OCR) and handwritten text recognition (HTR), Bengali handwritten text recognition, and the impact of tokenization strategies in OCR for complex scripts.

\paragraph{Optical Character Recognition.}
Text recognition has evolved from traditional connectionist temporal classification (CTC)$-$based models to more sophisticated sequence-to-sequence (seq2seq) approaches. Early CTC frameworks, such as CRNN \citep{shi2016end2end}, combined CNN-based visual feature extraction with recurrent networks for temporal modeling. Subsequent methods expanded on this by incorporating attention modules, alternative visual encoders, and advanced normalization techniques to address irregular text shapes and challenging image conditions \citep{gao2019reading, shi2018aster}.

The introduction of transformer architectures \citep{vaswani2017attention} led to significant advances in seq2seq modeling. Transformers, with their self-attention mechanisms and positional encodings, handle variations in character scale, spatial arrangements, and bidirectional decoding more effectively \citep{zhang2020sahan, lee2020recognizing}.

A more recent direction involves integrating language models (LMs) into recognition pipelines to enhance contextual understanding. TrOCR \citep{li2023trocr}, for example, leverages pretrained LMs as decoders using masked language modeling (MLM) objectives, while models like MaskOCR \citep{lyu2022maskocr} and ABINet \citep{fang2021read} explore more sophisticated strategies for linguistic pre-training. DTrOCR~\citep{fujitake2024dtrocr} advances this line of work by omitting the encoder entirely and relying on an autoregressively pretrained decoder (GPT-2) \citep{radford2019language}.

In the domain of handwritten text recognition (HTR), these trends are paralleled by augmenting traditional RNN+CTC architectures with attention and language modeling components \citep{memon2020handwritten, michael2019evaluating}.

\paragraph{Bengali Handwritten Text Recognition.}

Most existing research on Bengali handwritten text recognition has focused on constrained tasks, such as recognizing isolated characters or digits \citep{sufian2022bdnet}. A few works have extended this to word-level recognition using CRNN-based models trained with the CTC loss \citep{hossain2022lilaboti, basher2023bngraphemizer}. However, despite the growing prominence of transformer-based architectures in other languages, such methods remain largely unexplored for Bengali HTR.

\paragraph{Impact of Tokenization in OCR.}

Bengali script comprises approximately 13,000 graphemes~\citep{alam2021large}—substantially more than English, which has around 250. This makes the consideration of graphemes in tokenization a particularly important design decision in Bengali OCR. \citet{basher2023bngraphemizer} have shown empirically that grapheme-level tokenization yields improved performance in HTR compared to conventional character-level tokenization for Bengali.

\begin{figure*}[t]
  \centering
  \vspace{-3mm}
  \includegraphics[width=0.70\linewidth]{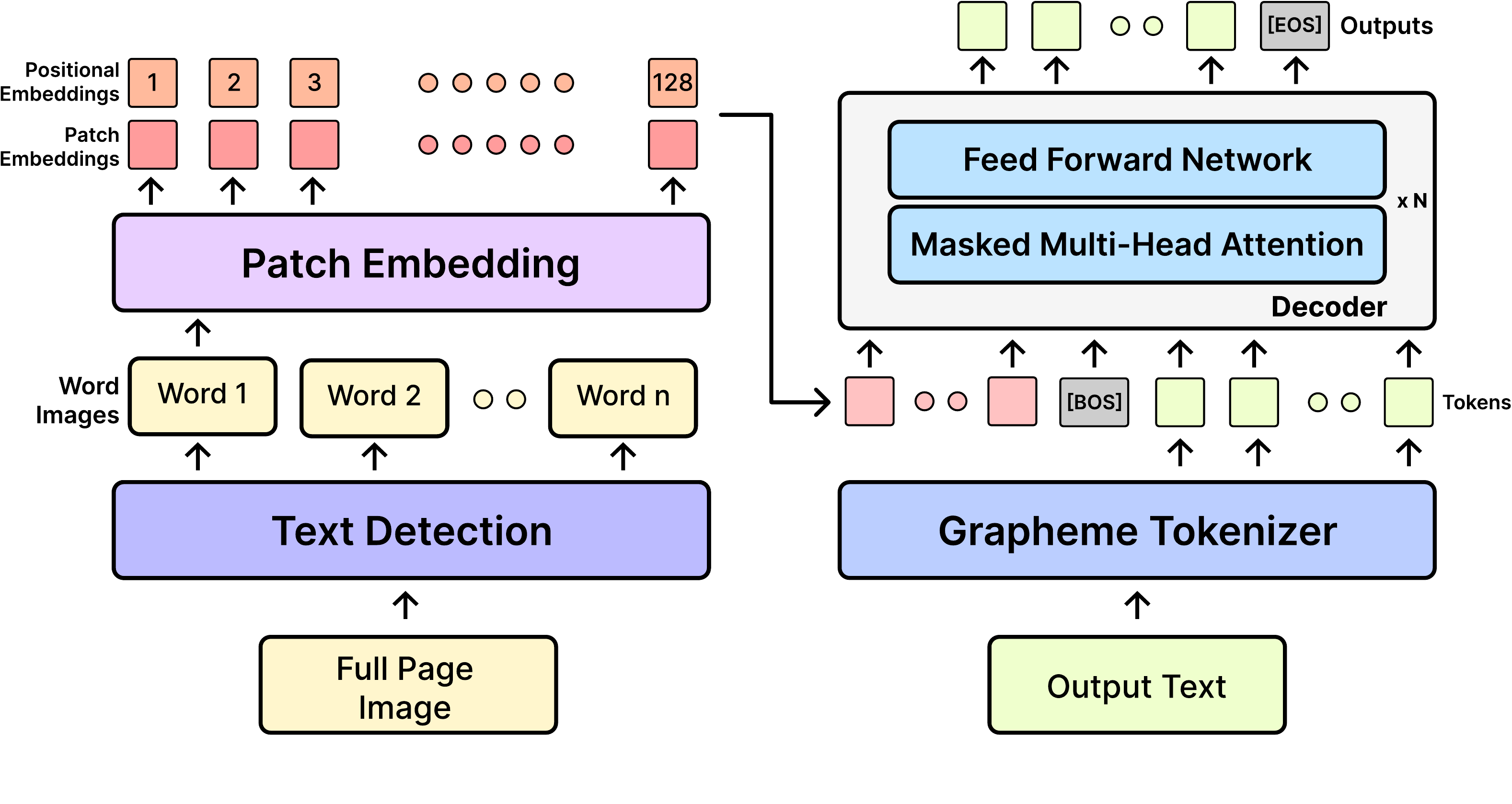}
   \vspace{-3mm}
  \caption{Architecture of the system pipeline. The pipeline consists of a text detection module and a text recognition module, which includes a patch embedding layer, a decoder-only transformer, and a grapheme-based tokenizer.}
  \label{fig:pipeline}
\end{figure*}
\section{Methodology}
The system pipeline consists of two primary components: a text detection module, which segments full-page images into individual word images, and a text recognition module, which adopts a decoder-only architecture, beginning with a patch embedding layer followed by a decoder-only transformer. To enhance recognition accuracy for Bengali script, a grapheme-based tokenizer is integrated into the decoder. The overall architecture of the system is illustrated in Figure~\ref{fig:pipeline}.

\subsection{Text Detection}
\label{sec:text_detection}

% As noted in Section~\ref{sec:introduction}, recognizing handwritten text from full-page images typically requires large annotated datasets, which are often unavailable for Bengali. Consequently, most HTR methods operate at the line or word level. Given the constraints of our data setting, we adopt a word-level recognition approach.

The task of recognizing handwritten text from full-page images typically requires large annotated datasets, which are often unavailable for Bengali. Consequently, most HTR methods operate at the line or word level. Given the constraints of our data setting, we adopt a word-level recognition approach.

% As illustrated in the lower-left portion of Figure~\ref{fig:pipeline}, the system begins by applying a text detection module to segment the full-page image into individual word images. For this, we employ BN-DRISHTI~\citep{jubaer2023bn}, a publicly available text detection model based on the YOLO framework~\citep{redmon2016yolo}. BN-DRISHTI performs a two-stage line segmentation process: an initial pass applies skew correction using Hough and Affine transformations, followed by refined line segmentation. Word images are then extracted from the segmented lines and passed to the recognition module.

% The system begins by applying a text detection module to segment the full-page image into individual word images. For this, we employ BN-DRISHTI~\citep{jubaer2023bn}, a publicly available text detection model based on the YOLO framework~\citep{redmon2016yolo}. BN-DRISHTI performs a two-stage line segmentation process: an initial line segmentation with skew correction using Hough and Affine transformations, followed by refined line segmentation. Word images are then extracted from the segmented lines and passed to the recognition module.

The system begins by applying a text detection module to segment the full-page image into individual word images. For this, we employ BN-DRISHTI~\citep{jubaer2023bn}, a publicly available text detection model based on the YOLO framework~\citep{redmon2016yolo}. 

BN-DRISHTI follows a two-step line segmentation process. In the first step, line regions are identified directly from the full-page image. However, due to the curvilinear nature of Bengali handwriting, these preliminary detections often include spurious boundaries or lines above or below the actual lines. To correct for this, Hough line and Affine transformations are applied to normalize skew and straighten the text baselines. Preprocessing operations such as binarization and morphological filtering are then employed to suppress noise and refine the line boundaries. In the second step, the refined line images are re-segmented, yielding more accurate line-level outputs. Finally, individual word images are segmented from the refined text lines and forwarded to the recognition module.\footnote{Since we employ the exact model without any modification, we refer the interested readers to the BN-DRISHTI paper~\citep{jubaer2023bn} for detailed methodology and evaluation results of the detection model.}

% According to the original paper, evaluation of the model on the test partition of the BN-HTRd~\citep{jubaer2023bn} dataset yields an F-score of 99.97\% for line segmentation and 98\% for word segmentation.

%Recognizing handwritten text from full-page images requires large annotated datasets, which are often unavailable. As a result, most HTR methods operate at the line or word level. Given our limited data setting, we adopt a word-level recognition approach.

%To segment full-page images into individual word images, we use BN-DRISHTI~\citep{jubaer2023bn}, a publicly available YOLO~\citep{redmon2016yolo} based text detection model. It performs a two-stage line segmentation: an initial pass with skew correction using Hough and Affine transformations, followed by refined segmentation. Word images are then segmented from the final segmented lines and passed to the recognition module.

\subsection{Text Recognition}
\label{sec:text_recognition}

We adopt a decoder-only text recognition module, in contrast to the encoder-decoder architecture commonly used in OCR. This module builds upon the open-source implementation of DTrOCR\footnote{\url{https://github.com/arvindrajan92/DTrOCR}}~\citep{fujitake2024dtrocr}, which we adapt using a grapheme-based tokenizer tailored for Bengali handwriting.

\subsubsection{Patch Embedding}

Transformers cannot directly process raw image inputs. Following the approach of ViT~\citep{dosovitskiy2021image}, each input image of size $(3, H_0, W_0)$ is first resized to $(3, H, W)$ and then divided into non-overlapping patches of size $(p_h, p_w)$, resulting in $N = (H / p_h) \times (W / p_w)$ patches. Each patch is linearly projected into a $D$-dimensional embedding. Positional encodings $P \in \mathbb{R}^{N \times D}$ are then added to retain spatial information. The resulting sequence is passed to the decoder module for autoregressive token prediction.

\subsubsection{Decoder}

The decoder module generates text directly from the embedded image sequence, bypassing the encoder used in traditional encoder–decoder architectures to extract visual features from raw images. This design eliminates the need for cross-attention and reduces both model parameters and computational cost. Our decoder module uses the architecture of Generative Pretrained Transformers (GPT)~\citep{radford2018improving, radford2019language} to recognize handwritten text. Specifically, we adopt a decoder-only architecture that consists solely of standard Transformer decoder blocks~\citep{vaswani2017attention}. In contrast to models such as TrOCR~\citep{li2023trocr} and DTrOCR~\citep{fujitake2024dtrocr}, we do not rely on pretrained Bengali language models. Instead, our decoder is trained from scratch, making it suitable for low-resource settings.

% [TODO: reviewer: write something on gpt-2?]

% Our recognition model uses only standard transformer decoder blocks~\citep{vaswani2017attention}, without an encoder. This eliminates the need for cross-attention and reduces both model parameters and computational cost. Each decoder block consists of masked multi-head self-attention and a feed-forward network. In contrast to models like TrOCR~\citep{li2023trocr} and DTrOCR~\citep{fujitake2024dtrocr}, we do not rely on pretrained Bengali language models. Instead, our decoder is trained from scratch, making it suitable for low-resource settings.

% After patch embedding, a \texttt{[BOS]} token is prepended to the sequence, which is then passed into the decoder. Tokens are generated autoregressively until a \texttt{[EOS]} token is produced. The final output from the decoder is projected to the vocabulary size and passed through a softmax layer for token prediction.

After the embedded image sequence, a \texttt{[BOS]} token is appended and the resulting sequence is passed into the decoder. Tokens are then generated autoregressively until a \texttt{[EOS]} token is produced. The final output from the decoder is projected onto the vocabulary space and passed through a softmax layer for token prediction.

\subsubsection{Tokenizer}

\begin{figure}[H]
  \centering
  \captionsetup[subfigure]{justification=centering, skip=1pt}

  \begin{subfigure}{\linewidth}
    \centering
    \includegraphics[width=0.85\linewidth]{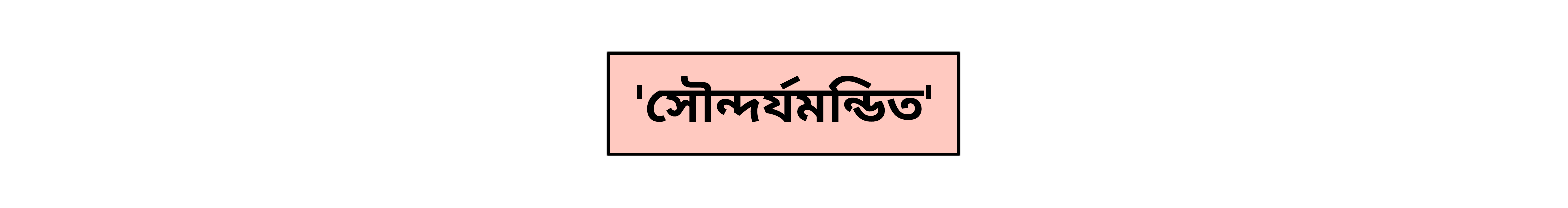}
    \caption{Example word.}
  \end{subfigure}

  \begin{subfigure}{\linewidth}
    \centering
    \includegraphics[width=0.85\linewidth]{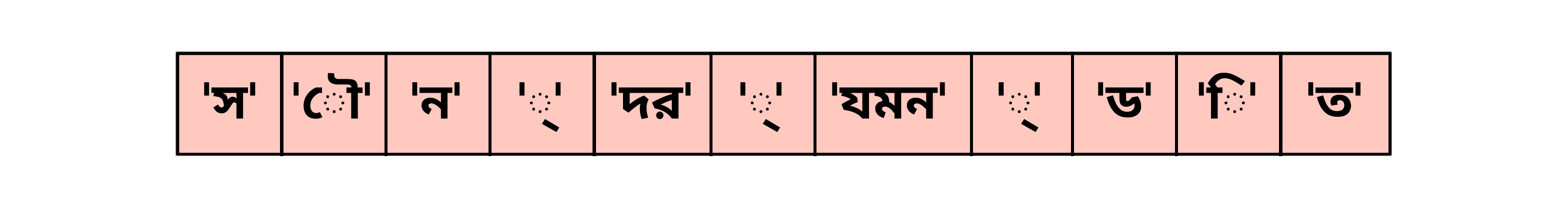}
    \caption{Byte pair encoding (BPE) tokenization.}
    \label{fig:bpe}
  \end{subfigure}

  \begin{subfigure}{\linewidth}
    \centering
    \includegraphics[width=0.85\linewidth]{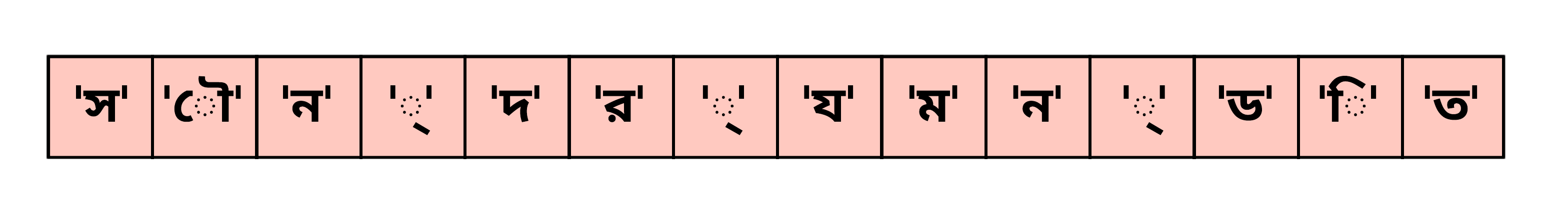}
    \caption{Character-level tokenization.}
    \label{fig:ct}
  \end{subfigure}

  \begin{subfigure}{\linewidth}
    \centering
    \includegraphics[width=0.85\linewidth]{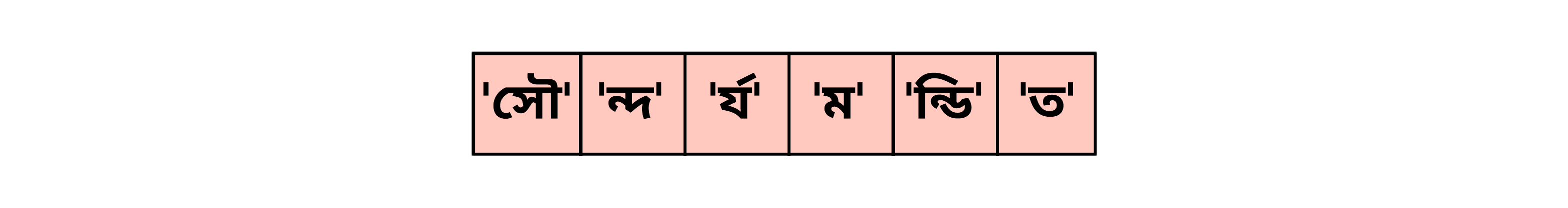}
    \caption{Grapheme-based tokenization.}
    \label{fig:grapheme}
  \end{subfigure}

  \caption{Comparison of tokenization strategies for Bengali HTR using the example word shown in (a).}

  \label{fig:tokenizations}
\end{figure}

A \textit{grapheme} is the smallest visually distinguishable unit in a writing system. Transformer-based OCR models~\citep{li2023trocr, fujitake2024dtrocr} typically rely on default tokenizers such as Byte Pair Encoding (BPE)~\citep{sennrich2016neural}, which segments text into subwords or characters based on frequency statistics (see Figure~\ref{fig:bpe}). These subword units often span multiple characters and are visually unstable in handwritten text: even slight variations in a single character can distort the entire token.

This challenge is amplified in Bengali, where writing styles vary widely and many graphemes are rendered with ligatures or conjuncts. Subword or character-level tokenization becomes unreliable without access to large, diverse training data. Furthermore, the visual order of Bengali text often diverges from its logical sequence, complicating alignment (see Figure~\ref{fig:ct}). Prior work~\citep{basher2023bngraphemizer} has shown that character-level tokenization fails to capture these structures effectively.

To address these challenges, we adopt a grapheme-based tokenizer that enforces a one-to-one mapping between visually coherent units and model output tokens (see Figure~\ref{fig:tokenizations}d). Specifically, we integrate BnGraphemizer~\citep{basher2023bngraphemizer}, a trie-based grapheme tokenizer designed for Bengali, into our decoder-only transformer pipeline.

\subsection{Training}

The training process consists of two stages: pre-training on large-scale synthetic data and fine-tuning on real-world, human-annotated samples. 

First, we employ a synthetic data generation tool\footnote{\url{https://github.com/tahsinchoudhury/GlyphScribe}} adapted from the VRD framework~\citep{lauar2024spanishtrocrleveragingtransfer} to construct diverse datasets (as detailed in Section~\ref{subsec:experimental_setup}) for pre-training. Pre-training itself is conducted in two phases, first on line-level images and then on word-level images. In the initial phase, the training on synthetic images containing single lines of handwritten text helps the model learn high-level structure. In the latter phase, the model is further trained on word-level images. Line-level training is particularly useful in early stages, as text detection models may group multiple words into a single detection region, especially in cursive or compact writing styles.

Second, we fine-tune the pretrained model on human-annotated Bengali handwriting datasets (as detailed in Section ~\ref{subsec:experimental_setup}), as previous works~\citep{baek2021what, bautista2022scene} have shown that synthetic-only dataset is generally insufficient for generalization to real-world inputs—an issue heightened in Bengali's complex script.

\section{Experiments}

In this section, we detail the experimental setup, describe the evaluation metrics, and present the results of our study.
% \subsection{Experimental Settings}

%  - list of metrics and state of the art methods and why ? 

\begin{table*}[ht]
\centering
\small
\begin{tabular}{@{}lccc cc cc@{}}
\toprule
\multirow{2}{*}{\textbf{Model}} 
  & \multirow{2}{*}{\textbf{Tokenizer}} 
  & \multirow{2}{*}{\shortstack{\textbf{Pretrained} \\ \textbf{LLM}}} 
  & \multirow{2}{*}{\textbf{\#Parameters}} 
  & \multicolumn{2}{c}{\textbf{BN-HTRd}} 
  & \multicolumn{2}{c}{\textbf{Bongabdo}} \\
\cmidrule(l){5-6}\cmidrule(l){7-8}
  &  &  &  & \textbf{CER(\%)} & \textbf{WER(\%)} & \textbf{CER(\%)} & \textbf{WER(\%)} \\
\midrule
Line\_BPE\_Pretrained          & BPE                    & \cmark & 162M & 38.34 & 51.83 & 65.04 & 77.65 \\
Line\_BPE\_Random          & BPE                    & \xmark  & 162M & 26.17 & 39.22 & 46.91 & 63.85 \\
Line\_BnG\_Random          & BnGraphemizer & \xmark  & 87M & 29.58 & 42.89 & 50.69 & 68.02 \\
Word\_BPE\_Pretrained          & BPE                    & \cmark & 162M &  7.57 & 16.30 & 11.61 & 26.99 \\
Word\_BPE\_Random          & BPE                    & \xmark  & 162M &  6.68 & 15.05 & 10.09 & 25.39 \\
\textbf{Word\_BnG\_Random} & BnGraphemizer & \xmark  & 87M & \textbf{6.19} & \textbf{14.20} & \textbf{8.68} & \textbf{23.56} \\
\bottomrule
\end{tabular}
\caption{Model configuration details and recognition performance (CER, WER) on the BN-HTRd and Bongabdo datasets. 
Model names follow the format \textit{[Input granularity]\_[Tokenizer]\_[Decoder initialization]}. Here, \textit{Line} level models take 
entire handwritten line images as input and predict the full line text, while \textit{Word} level models take cropped 
word images as input and predict the word text. \textit{BPE} and \textit{BnG} denote the Byte-Pair Encoding 
and BnGraphemizer tokenizers, respectively. \textit{Pretrained}/\textit{Random} indicates whether the decoder was initialized from a 
pretrained Bengali language model or with random weights. Among all variants, \textbf{Word\_BnG\_Random} (our proposed 
word-level model with grapheme tokenizer, trained from scratch) achieves the lowest CER and WER.}

\label{tab:model-config}
\end{table*}
\vspace{-2mm}

\subsection{Experimental Setup}
\label{subsec:experimental_setup}
We outline the experimental setup here, covering the details of training data and evaluation benchmarks, training configurations, and implementation details used in both the pre-training and fine-tuning stages.

% \subsubsection{Datasets}
\subsubsection{Training Data and Benchmarks}
For pre-training, we generate synthetic datasets comprising 4.5 million line-level samples and 7 million word-level samples. The text content is sourced from Bengali Wikipedia, the Bangla-NMT dataset~\citep{hasan2020nmt}, and a publicly available Bengali dictionary~\citep{kamalbengalidictionary}. These sources provide a wide range of vocabulary and writing styles. To enhance realism, our synthetic data generator introduces artifacts such as handwritten-style fonts, wavy or bent lines, Gaussian blur, and partial character fragments that mimic cropping effects. Representative examples are illustrated in Appendix~\ref{app:synthimages}.

To fine-tune and evaluate our model, we use six human-annotated datasets: BN-HTRd~\citep{rahman2023bnhtrd, jubaer2023bn}, BanglaWriting~\citep{mridha2021banglawriting}, BanglaLekha-Isolated~\citep{biswas2017banglalekha}, IIIT-Indic~\citep{jindal2021iiitindic}, ICDAR 2024~\citep{icdar2024dataset}, and Bongabdo~\citep{islam2023towards}. 

\begin{table}[H]
    \centering
    \small
    \begin{tabular}{l r}
        \toprule
        \textbf{Dataset} & \textbf{\# Words} \\
        \midrule
        BN-HTRd                & 104,854 \\
        BanglaWriting          & 21,234 \\
        BanglaLekha-Isolated (Digits only)  & 19,748 \\
        IIIT-Indic             & 113,075 \\
        ICDAR 2024             & 79,663 \\
        \midrule
        \textbf{Total} & \textbf{338,574} \\
        \bottomrule
    \end{tabular}
    \caption{Summary of human-labeled datasets used to fine-tune the word-level models.}
    \label{tab:dataset_statistics}
\end{table}

We consider two model variants: line-level models, which take entire handwritten line images as input and predict the full line text, and word-level models, which take cropped word images as input and predict the text for that word. For fine-tuning, the line models use 13,912 real handwritten line images from the BN-HTRd dataset~\citep{rahman2023bnhtrd}. In contrast, the word models are fine-tuned on a larger collection of approximately 340,000 (see Table~\ref{tab:dataset_statistics} for details) real handwritten word images aggregated from BN-HTRd~\citep{rahman2023bnhtrd}, BanglaWriting, BanglaLekha-Isolated, IIIT-Indic, and ICDAR 2024. For evaluation, we benchmark on 805 full-page images from the “automatic annotation” split of the extended BN-HTRd dataset~\citep{jubaer2023bn} and 111 images from the Bongabdo dataset.

\subsubsection{Implementation Details}
We implement our decoder using the GPT-2 architecture from HuggingFace~\citep{wolf2020huggingface}, consisting of 12 transformer layers with a hidden size of 768 and 12 self-attention heads. The maximum sequence length is set to 256 tokens. To better suit the properties of Bengali handwritten text, we replace the default byte-pair encoding tokenizer with the BnGraphemizer tokenizer.

Each input image is resized to a fixed resolution of $32 \times 128$ pixels and divided into non-overlapping patches of size $4 \times 8$, yielding 128 image tokens per word. The model is pretrained using a batch size of 32 with the Adam optimizer and a learning rate of $1 \times 10^{-4}$. During fine-tuning, the same batch size and optimizer are used, but the learning rate is reduced to $5 \times 10^{-6}$. 

All training and inference experiments are conducted using PyTorch on a single NVIDIA GeForce RTX 3080 GPU (see Appendix~\ref{app:traindetails} for more details).

% \subsubsection{Baselines}
% We benchmark our model against two categories of baselines: existing Bengali OCR systems such as Gemini 2.5 Flash~\citep{comanici2025gemini} and Imagetotext.info\footnote{https://www.imagetotext.info}, and prior Bengali HTR models including LILA-BOTI~\citep{hossain2022lilaboti}, CRNN-CT, and CRNN-BnG~\citep{basher2023bngraphemizer}.

%  Dataset 
\subsection{Evaluation Metrics}

We evaluate our handwriting recognition system using two widely adopted metrics for handwritten text recognition (HTR): Character Error Rate (CER) and Word Error Rate (WER). CER measures the proportion of character-level errors—substitutions, deletions, and insertions—required to transform the predicted sequence into the ground truth, while WER computes the same at the word level. Formal definitions of both metrics are provided in Appendix~\ref{app:evalmetrics}.

\subsection{Experimental Results}
\label{sec:evaluation}

We now discuss the results of our model across ablations, comparisons with existing OCR systems and prior HTR models, an analysis of its sensitivity to text detection quality and inference speed.

\paragraph{Model Configuration Analysis.}

We compare multiple model configurations to evaluate the effects of input granularity, decoder initialization, and tokenization. Specifically, we study line-level vs.\ word-level recognition, pretrained Bengali language model vs.\ randomly initialized decoders, and subword-level Byte Pair Encoding (BPE) vs.\ the grapheme-level BnGraphemizer tokenizer. While BPE is used with the pretrained LLM, BnGraphemizer is evaluated only with randomly initialized decoders, as no pretrained Bengali LLM exists for this tokenizer.

The line-level models are pretrained on synthetic line images, while word-level models undergo additional pretraining on synthetic word images. Line-level models are fine-tuned on handwritten line images, while word-level models are fine-tuned on a larger set of handwritten word images, as discussed in Section~\ref{subsec:experimental_setup}.
% For fine-tuning, the line models use 13,912 real handwritten line images from the BN-HTRd dataset. In contrast, the word models are fine-tuned on a larger dataset of approximately 330,000 real handwritten word images (see Appendix~\ref{app:real_data} for details). All models are evaluated on 805 full-page images from the “automatic annotation” split of the extended BN-HTRd dataset~\citep{jubaer2023bn} and 111 images from the Bongabdo dataset.

As shown in Table~\ref{tab:model-config}, word-level models outperform line-level ones, which can be attributed to the abundance of word-level training data and the more nuanced nature of line-level data. Our proposed model—Word\_BnG\_Random—achieves the lowest CER and WER without using a pretrained Bengali LLM. The grapheme-based tokenizer further improves accuracy while reducing model size from 162M to 87M, demonstrating both effectiveness and parameter efficiency.

\begin{table}[t]
\centering
\small
\resizebox{\columnwidth}{!}{%
\begin{tabular}{@{}lccrr@{}}
\toprule
\textbf{System}       & \multicolumn{2}{c}{\textbf{BN-HTRd}} & \multicolumn{2}{c}{\textbf{Bongabdo}} \\
\cmidrule(lr){2-3}\cmidrule(lr){4-5}
                     & \textbf{CER (\%)} & \textbf{WER (\%)} & \textbf{CER (\%)} & \textbf{WER (\%)} \\
\midrule
Gemini 2.5 Flash\tablefootnote{Although Gemini 2.5 Pro yields slightly better results, we use Flash due to strict API limits in the Pro free tier.}  
                     & 19.39             & 32.49             & 37.42             & 56.39             \\
Imagetotext.info\tablefootnote{\url{https://www.imagetotext.info}}     & 20.90             & 37.10             & 14.36             & 31.14             \\
\textbf{Ours}        & \textbf{6.19}     & \textbf{14.20}    & \textbf{8.68}     & \textbf{23.56}    \\
\bottomrule
\end{tabular}%
}
\caption{Comparison of cloud-based OCR systems and our proposed model on Bengali HTR.}
\label{tab:ocr_services_comparison}
\end{table}

% \footnotetext[1]{Although Gemini 2.5 Pro yields slightly better results, we use the Flash due to strict API limits in the Pro free tier.}
% \footnotetext[2]{\url{https://www.imagetotext.info}}

\paragraph{Comparison with Existing OCR Systems.}
We compare our proposed model with two existing OCR systems: Gemini 2.5 Flash and \href{https://www.imagetotext.info/}{Imagetotext.info}. As shown in Table~\ref{tab:ocr_services_comparison}, our model achieves the lowest CER and WER, outperforming both systems by a large margin. Gemini 2.5 Flash is selected as a representative vision-language model (see Appendix~\ref{app:llmeval} for prompt details) due to its strong performance on Bengali handwritten text, substantially surpassing models such as ChatGPT and Claude on this task. We exclude other Google Vision API models, as Gemini consistently outperforms them. \href{https://www.imagetotext.info/}{Imagetotext.info} is included for its support for Bengali handwriting and its accessible API, which enables automated benchmarking. Other OCR services were excluded based on lack of Bengali handwriting support, absence of APIs, or poor performance in preliminary evaluations.

%\paragraph{Comparison with Existing OCR Systems.}
%We compare our proposed model against Gemini 2.5 Flash and Imagetotext.info to evaluate its performance against existing OCR systems.  Table~\ref{tab:ocr_services_comparison} shows that our model achieves the lowest CER and WER, outperforming both systems by a large margin. Gemini 2.5 Flash was selected as a representative vision-language model because it performs significantly better on Bengali handwritten text recognition than other models like ChatGPT or Claude. We did not include other Google Vision API models, as Gemini significantly outperforms them on this task. Imagetotext.info was chosen for its Bengali handwriting support and accessible API, which allowed automated benchmarking. Other online OCR services were excluded due to lack of Bengali handwriting recognition, absence of APIs, or poor recognition performance in preliminary testing.

\begin{table}[t]
\centering
\small
\resizebox{\columnwidth}{!}{%
\begin{tabular}{@{}lcccc@{}}
\toprule
\textbf{Filtered} & \multicolumn{2}{c}{\textbf{BN-HTRd}} & \multicolumn{2}{c}{\textbf{Bongabdo}} \\
\cmidrule(lr){2-3}\cmidrule(lr){4-5}
                  & \textbf{CER (\%)} & \textbf{WER (\%)} & \textbf{CER (\%)} & \textbf{WER (\%)} \\
\midrule
No               & 6.19              & 14.20             & 8.68              & 23.56             \\
Yes                & \textbf{4.58}     & \textbf{12.59}    & \textbf{8.02}     & \textbf{22.95}    \\
\bottomrule
\end{tabular}%
}
\caption{Comparison of recognition performance with and without filtering images where the text detection module failed to segment words accurately.}
\label{tab:filtered_results}
\end{table}

\paragraph{Dependency on Text Detection Model.}
To assess the sensitivity of our system to the quality of word segmentation, we manually inspected the BN-HTRd and Bongabdo datasets and identified a small number of samples where the text detection model failed to accurately segment words. After filtering out these cases, our text recognition model demonstrated improved performance, as shown in Table~\ref{tab:filtered_results}. These findings suggest that the end-to-end pipeline is affected by errors in text detection, and that further gains in recognition accuracy may be achieved by incorporating a more precise word segmentation module.

%\paragraph{Dependency on Text Detection Model.}
%We manually inspected the BN-HTRd and Bongabdo datasets and identified a small number of images where the text detection model failed to accurately segment text into individual words. After filtering out these samples, our text recognition model showed improved performance (Table~\ref{tab:filtered_results}). This result suggests that the overall pipeline is sensitive to the quality of word segmentation, and that a more reliable text detection model could further enhance recognition accuracy.

\begin{table}[t]
\centering
\small
\begin{tabular}{@{}llcc@{}}
\toprule
\textbf{Datasets} & \textbf{Model} & \textbf{CER (\%)} & \textbf{WER (\%)} \\
\midrule
\multirow{4}{*}{\makecell[l]{BW (train)\\BN-HTRd (test)}} 
  & LILA-BOTI$^{*}$     & 25.92 & 51.62 \\
  & CRNN-CT$^{*}$                  & 19.94 & 44.10 \\
  & CRNN-BnG$^{*}$         & 30.53 & 48.41 \\
  & \textbf{Ours}            & \textbf{18.04} & \textbf{33.20} \\
\midrule
\multirow{4}{*}{\makecell[l]{BN-HTRd (train)\\BW (test)}} 
  & LILA-BOTI$^{*}$     & 15.54 & 36.78 \\
  & CRNN-CT$^{*}$                  & \textbf{11.12} & 29.43 \\
  & CRNN-BnG$^{*}$          & 15.80 & 29.18 \\
  & \textbf{Ours}            & 12.66 & \textbf{24.79} \\
\bottomrule
\end{tabular}
\caption{Cross-dataset evaluation on BW (BanglaWriting) and BN-HTRd datasets. The asterisk ($^{*}$) indicates results reported from~\citep{basher2023bngraphemizer}.}
\label{tab:cross_dataset}
\end{table}

\paragraph{Comparison with Prior Models.}

Table~\ref{tab:cross_dataset} presents cross-dataset evaluation results comparing our proposed model with prior Bengali HTR approaches, including LILA-BOTI~\citep{hossain2022lilaboti}, CRNN-CT, and CRNN-BnG~\citep{basher2023bngraphemizer}. We adopt cross-dataset evaluation to ensure consistency with prior work, which report performance under the same setting. This approach provides a fair basis for comparison and reflects the models’ ability to generalize across different handwriting corpora. When fine-tuned on BW (BanglaWriting)~\citep{mridha2021banglawriting} and evaluated on BN-HTRd~\citep{rahman2023bnhtrd}, our model achieves the lowest CER and WER. In the reverse setting, it obtains the best WER, while the CER is slightly higher. 

\paragraph{Inference Speed.}

We conducted experiments on the Bongabdo dataset to compare the inference speeds of different models, as reported in Table~\ref{tab:inference_bongabdo}. Each experiment was repeated five times to ensure statistical robustness, and the average inference speed was reported as the final result. This set of experiments was performed on the aforementioned devices, which were also used for model training and other experiments. As expected, our proposed model with 87M parameters achieves a higher inference speed than the 162M parameter variant. This demonstrates the practicality of our model for deployment in low-resource computational settings while delivering faster results to end users.

\begin{table}[H]
\centering
\small
\resizebox{\columnwidth}{!}{%
\begin{tabular}{@{}lcccc@{}}
\toprule
\textbf{Model} & \textbf{\#Params} & \textbf{\#Words} & \textbf{Time} & \textbf{Speed} \\
\midrule
Word\_BPE\_Random & 162M & 17,217 & 2074.32s & 8.30 words/s \\
\textbf{Word\_BnG\_Random} & 87M  & 17,217 & \textbf{1729.59s} & \textbf{9.95} words/s \\
\bottomrule
\end{tabular}%
}
\caption{Inference time on the Bongabdo dataset.}
\label{tab:inference_bongabdo}
\end{table}

\section{System Description}
\vspace{-1.5mm}
We deploy our Bengali handwriting recognition system as a Flask-based REST API, enabling external access through a lightweight and modular interface. The backend supports asynchronous task execution using Celery, with Redis serving as the message broker. To ensure reproducibility and ease of deployment, the entire system is containerized with Docker, and all components are orchestrated via Docker Compose. The inference engine runs on an NVIDIA GeForce RTX 3080 GPU.

% We deploy our Bengali handwriting recognition system as a Flask-based REST API, enabling external access through a lightweight and modular interface. The backend supports asynchronous task execution using Celery, with Redis serving as the message broker. To ensure reproducibility and ease of deployment, the entire system is containerized with Docker, and all components—including the Flask API server, Celery workers, and Redis—are orchestrated via Docker Compose. The inference engine runs on an NVIDIA GeForce RTX 3080 GPU.
% , providing sufficient compute for low-latency operation.

The system supports a variety of input formats, including individual image files (PNG, JPG, JPEG) as well as multi-page PDF documents. Once uploaded, documents are processed through the detection and recognition pipeline described in Sections~\ref{sec:text_detection} and~\ref{sec:text_recognition}, respectively, where full-page handwritten images are segmented into word-level regions and transcribed using a decoder-only transformer model. For each image or PDF page, the recognized text is displayed in an editable text area, allowing users to make corrections as needed. In the case of PDFs, the system provides per-page navigation to support multi-page review. Users can export the extracted text in either plain text (\texttt{.txt}) or Microsoft Word (\texttt{.docx}) format.

The web interface features a document upload panel on the left and a text output display area on the right. Snapshots of the user interface, along with representative use cases, are provided in Appendix~\ref{app:sampleoutputs}.

\section{Conclusions}
We present a resource-efficient Bengali handwritten text recognition system that operates at the word level without relying on a pretrained large language model as the decoder. Our approach leverages a decoder-only transformer combined with a grapheme-based tokenizer tailored for Bangla script. Experimental results demonstrate that our method outperforms existing OCR systems and prior Bengali HTR models, achieving state-of-the-art accuracy despite being trained on limited annotated data.

% \paragraph{Future Work.}
% Write future work.

\section{Limitations and Future Work}
Our system has two main limitations: its training data mainly consists of handwritten text on white backgrounds, limiting adaptability to more varied real-world scenarios; and the scarcity of large-scale pretrained GPT-2 models for Bengali limits the breadth of the experiments. Future work can include expanding the synthetic dataset to cover diverse backgrounds and noise artifacts, pre-training auto-regressive language models on extensive Bengali text, experimenting with other auto-regressive language models (such as Llama), and refining text detection to reduce segmentation errors. To support future research in Bengali HTR, we publicly release our complete pipeline under the MIT license.

\section*{Ethical Considerations}
Our system does not retain any user-submitted documents or images. All uploads are processed temporarily and permanently deleted after a short duration to ensure user privacy. The datasets used for training and evaluation are publicly available and intended for research purposes. No personal or sensitive content was included in the model development process.

We promote responsible use of the system in archival, educational, and administrative contexts. Its intended applications include the digitization of historical or administrative documents, support for literacy development, and automation of handwritten workflows in institutional settings. It is not designed for surveillance, profiling, or other harmful purposes.

\section*{Acknowledgements}
The authors acknowledge the support provided by the Bangladesh Bureau of Educational Information and Statistics (BANBEIS) and the University of Dhaka during the course of this research.

\bibliography{custom}

\begin{thebibliography}{41}
\providecommand{\natexlab}[1]{#1}

\bibitem[{Alam et~al.(2021)Alam, Reasat, Sushmit, Siddique, Rahman, Hasan, and Humayun}]{alam2021large}
S.~Alam, T.~Reasat, A.~S. Sushmit, S.~M. Siddique, F.~Rahman, M.~Hasan, and A.~I. Humayun. 2021.
\newblock A large multi-target dataset of common bengali handwritten graphemes.
\newblock In \emph{Document Analysis and Recognition – ICDAR 2021}, pages 383--398. Springer International Publishing.

\bibitem[{Azad et~al.(2020)Azad, Singha, and Nahid}]{azad2020bangla}
M.~A. Azad, H.~S. Singha, and M.~M.~H. Nahid. 2020.
\newblock Bangla handwritten character recognition using deep convolutional autoencoder neural network.
\newblock In \emph{2020 2nd International Conference on Advanced Information and Communication Technology (ICAICT)}, pages 295--300.

\bibitem[{Baek et~al.(2021)Baek, Matsui, and Aizawa}]{baek2021what}
Jeonghun Baek, Yusuke Matsui, and Kiyoharu Aizawa. 2021.
\newblock What if we only use real datasets for scene text recognition? toward scene text recognition with fewer labels.
\newblock In \emph{Proceedings of the IEEE/CVF Conference on Computer Vision and Pattern Recognition (CVPR)}, pages 3113--3122.

\bibitem[{Basher et~al.(2023)Basher, Noor, Afroze, and Ahmed}]{basher2023bngraphemizer}
Mohammad Jahid~Ibna Basher, Mohammad~Raghib Noor, Sadia Afroze, and Ikbal Ahmed. 2023.
\newblock \href {https://doi.org/10.1109/WIECON-ECE60392.2023.10456463} {Bngraphemizer: A grapheme-based tokenizer for bengali handwritten text recognition}.
\newblock In \emph{2023 IEEE International Women in Engineering (WIE) Conference on Electrical and Computer Engineering}, Sree Chitra Thirunal College of Engineering, Thiruvananthapuram, Kerala, INDIA.

\bibitem[{Bautista and Atienza(2022)}]{bautista2022scene}
Darwin Bautista and Rowel Atienza. 2022.
\newblock Scene text recognition with permuted autoregressive sequence models.
\newblock In \emph{Proceedings of the European Conference on Computer Vision (ECCV)}, pages 178--196.

\bibitem[{Biswas et~al.(2017)Biswas, Islam, Shom, Shopon, Mohammed, Momen, and Abedin}]{biswas2017banglalekha}
Mithun Biswas, Rafiqul Islam, Gautam~Kumar Shom, Md. Shopon, Nabeel Mohammed, Sifat Momen, and Anowarul Abedin. 2017.
\newblock \href {https://doi.org/10.1016/j.dib.2017.03.045} {Banglalekha-isolated: A multi-purpose comprehensive dataset of handwritten bangla isolated characters}.
\newblock \emph{Data in Brief}, 12:103--107.

\bibitem[{Chaudhury et~al.(2022)Chaudhury, Mukherjee, Das, Biswas, and Bhattacharya}]{chaudhury2022deep}
A.~Chaudhury, P.~S. Mukherjee, S.~Das, C.~Biswas, and U.~Bhattacharya. 2022.
\newblock A deep ocr for degraded bangla documents.
\newblock \emph{ACM Transactions on Asian and Low-Resource Language Information Processing}, 21(5).

\bibitem[{Dosovitskiy et~al.(2021)Dosovitskiy, Beyer, Kolesnikov, Weissenborn, Zhai, Unterthiner, Dehghani, Minderer, Heigold, Gelly, Uszkoreit, and Houlsby}]{dosovitskiy2021image}
Alexey Dosovitskiy, Lucas Beyer, Alexander Kolesnikov, Dirk Weissenborn, Xiaohua Zhai, Thomas Unterthiner, Mostafa Dehghani, Matthias Minderer, Georg Heigold, Sylvain Gelly, Jakob Uszkoreit, and Neil Houlsby. 2021.
\newblock \href {https://openreview.net/forum?id=YicbFdNTTy} {An image is worth 16x16 words: Transformers for image recognition at scale}.
\newblock In \emph{Proceedings of the International Conference on Learning Representations (ICLR)}.

\bibitem[{Fang et~al.(2021)Fang, Xie, Wang, Mao, and Zhang}]{fang2021read}
Shancheng Fang, Hongtao Xie, Yuxin Wang, Zhendong Mao, and Yongdong Zhang. 2021.
\newblock Read like humans: Autonomous, bidirectional and iterative language modeling for scene text recognition.
\newblock In \emph{Proceedings of the IEEE/CVF Conference on Computer Vision and Pattern Recognition (CVPR)}, pages 7098--7107.

\bibitem[{Fujitake(2024)}]{fujitake2024dtrocr}
Masato Fujitake. 2024.
\newblock Dtrocr: Decoder-only transformer for optical character recognition.
\newblock In \emph{Proceedings of the IEEE/CVF Winter Conference on Applications of Computer Vision (WACV)}, pages 8025--8035.

\bibitem[{Gao et~al.(2019)Gao, Chen, Wang, Tang, and Lu}]{gao2019reading}
Y.~Gao, Y.~Chen, J.~Wang, M.~Tang, and H.~Lu. 2019.
\newblock Reading scene text with fully convolutional sequence modeling.
\newblock \emph{Neurocomputing}, 339:161--170.

\bibitem[{Graves et~al.(2006)Graves, Fernandez, Gomez, and Schmidhuber}]{graves2006}
A.~Graves, S.~Fernandez, F.~Gomez, and J.~Schmidhuber. 2006.
\newblock Connectionist temporal classification: labelling unsegmented sequence data with recurrent neural network.
\newblock In \emph{Proceedings of the 23rd international conference on Machine learning}.

\bibitem[{Hasan et~al.(2020)Hasan, Bhattacharjee, Samin, Hasan, Basak, Rahman, and Shahriyar}]{hasan2020nmt}
Tahmid Hasan, Abhik Bhattacharjee, Kazi Samin, Masum Hasan, Madhusudan Basak, M.~Sohel Rahman, and Rifat Shahriyar. 2020.
\newblock \href {https://doi.org/10.18653/v1/2020.emnlp-main.207} {Not low-resource anymore: Aligner ensembling, batch filtering, and new datasets for {B}engali-{E}nglish machine translation}.
\newblock In \emph{Proceedings of the 2020 Conference on Empirical Methods in Natural Language Processing (EMNLP)}, pages 2612--2623, Online. Association for Computational Linguistics.

\bibitem[{Hossain et~al.(2022)Hossain, Rakib, Mollah, Rahman, and Mohammed}]{hossain2022lilaboti}
M.~I. Hossain, M.~Rakib, S.~Mollah, F.~Rahman, and N.~Mohammed. 2022.
\newblock Lila-boti: Leveraging isolated letter accumulations by ordering teacher insights for bangla handwriting recognition.
\newblock In \emph{2022 26th International Conference on Pattern Recognition (ICPR)}, pages 1770--177.

\bibitem[{Hossain et~al.(2021)Hossain, Hasan, and Das}]{hossain2021bangla}
M.~T. Hossain, M.~W. Hasan, and A.~K. Das. 2021.
\newblock Bangla handwritten word recognition system using convolutional neural network.
\newblock In \emph{2021 15th International Conference on Ubiquitous Information Management and Communication (IMCOM)}, pages 1--8.

\bibitem[{{IIIT Hyderabad and CVIT Lab}(2024)}]{icdar2024dataset}
{IIIT Hyderabad and CVIT Lab}. 2024.
\newblock Icdar 2024 handwritten document recognition dataset.
\newblock \url{https://ilocr.iiit.ac.in/icdar_2024_hwd/}.

\bibitem[{Islam et~al.(2023)Islam, Das, Kowsar, Rabby, Hasan, and Rahman}]{islam2023towards}
Md~Majedul Islam, Avishek Das, Ibna Kowsar, AKM Shahariar~Azad Rabby, Nazmul Hasan, and Fuad Rahman. 2023.
\newblock \href {https://doi.org/10.1109/BigData59063.2023.10372570} {Towards full-page offline bangla handwritten text recognition using image-to-sequence architecture}.
\newblock In \emph{2023 IEEE International Conference on Big Data (Big Data)}, pages 1061--1067. IEEE.

\bibitem[{Jindal et~al.(2021)Jindal, Kumar, and Jawahar}]{jindal2021iiitindic}
Milan Jindal, Pratyush Kumar, and C.V. Jawahar. 2021.
\newblock \href {https://arxiv.org/abs/2105.04020} {iiit-indic-hw-words: A dataset for indic handwritten text recognition}.
\newblock \emph{arXiv preprint arXiv:2105.04020}.

\bibitem[{Jubaer et~al.(2023)Jubaer, Tabassum, Rahman, and Islam}]{jubaer2023bn}
Sheikh~Mohammad Jubaer, Nazifa Tabassum, Md~Ataur Rahman, and Mohammad~Khairul Islam. 2023.
\newblock Bn-drishti: Bangla document recognition through instance-level segmentation of handwritten text images.
\newblock \emph{arXiv preprint arXiv:2306.09351}.

\bibitem[{Kamal(2018)}]{kamalbengalidictionary}
M.~Kamal. 2018.
\newblock Minhaskamal/bengalidictionary: A large collection of bengali words.
\newblock \url{https://github.com/MinhasKamal/BengaliDictionary}.

\bibitem[{Lauar and Laurent(2024)}]{lauar2024spanishtrocrleveragingtransfer}
Filipe Lauar and Valentin Laurent. 2024.
\newblock \href {https://arxiv.org/abs/2407.06950} {Spanish trocr: Leveraging transfer learning for language adaptation}.
\newblock \emph{Preprint}, arXiv:2407.06950.

\bibitem[{Lee et~al.(2020)Lee, Park, Baek, Oh, Kim, and Lee}]{lee2020recognizing}
J.~Lee, S.~Park, J.~Baek, S.J. Oh, S.~Kim, and H.~Lee. 2020.
\newblock On recognizing texts of arbitrary shapes with 2d self-attention.
\newblock In \emph{Proceedings of the IEEE/CVF Conference on Computer Vision and Pattern Recognition Workshops}, pages 546--547.

\bibitem[{Li et~al.(2023)Li, Lv, Chen, Cui, Lu, Florencio, Zhang, Li, and Wei}]{li2023trocr}
Minghao Li, Tengchao Lv, Jingye Chen, Lei Cui, Yijuan Lu, Dinei Florencio, Cha Zhang, Zhoujun Li, and Furu Wei. 2023.
\newblock Trocr: Transformer-based optical character recognition with pre-trained models.
\newblock In \emph{Proceedings of the AAAI Conference on Artificial Intelligence}, pages 13094--13102.

\bibitem[{Lyu et~al.(2022)Lyu, Zhang, Liu, Qiao, Xu, Wu, Yao, Han, Ding, and Wang}]{lyu2022maskocr}
Hengyuan Lyu, Chengquan Zhang, Shanshan Liu, Meina Qiao, Yangliu Xu, Liang Wu, Kun Yao, Junyu Han, Errui Ding, and Jingdong Wang. 2022.
\newblock Maskocr: Text recognition with masked encoder-decoder pretraining.
\newblock \emph{arXiv preprint arXiv:2206.00311}.

\bibitem[{Memon et~al.(2020)Memon, Sami, Khan, and Uddin}]{memon2020handwritten}
J.~Memon, M.~Sami, R.~A. Khan, and M.~Uddin. 2020.
\newblock Handwritten optical character recognition (ocr): A comprehensive systematic literature review (slr).
\newblock \emph{IEEE Access}, 8:142642--142668.

\bibitem[{Michael et~al.(2019)Michael, Labahn, Grüning, and Zöllner}]{michael2019evaluating}
J.~Michael, R.~Labahn, T.~Grüning, and J.~Zöllner. 2019.
\newblock Evaluating sequence-to-sequence models for handwritten text recognition.
\newblock In \emph{2019 International Conference on Document Analysis and Recognition (ICDAR)}, pages 1286--1293. IEEE.

\bibitem[{Mridha et~al.(2021)Mridha, Ohi, Ali, Emon, and Kabir}]{mridha2021banglawriting}
Md~Forhad Mridha, Abu~Quwsar Ohi, Md~Ameer Ali, Md~I Emon, and Md~Mohsin Kabir. 2021.
\newblock \href {https://doi.org/10.1016/j.dib.2021.106633} {Banglawriting: A multi-purpose offline bangla handwriting dataset}.
\newblock \emph{Data in Brief}, 34:106633.

\bibitem[{Radford et~al.(2018)Radford, Narasimhan, Salimans, and Sutskever}]{radford2018improving}
Alec Radford, Karthik Narasimhan, Tim Salimans, and Ilya Sutskever. 2018.
\newblock \href {https://cdn.openai.com/research-covers/language-unsupervised/language_understanding_paper.pdf} {Improving language understanding by generative pre-training}.
\newblock \emph{OpenAI}.
\newblock Technical report.

\bibitem[{Radford et~al.(2019)Radford, Wu, Child, Luan, Amodei, and Sutskever}]{radford2019language}
Alec Radford, Jeffrey Wu, Rewon Child, David Luan, Dario Amodei, and Ilya Sutskever. 2019.
\newblock \href {https://cdn.openai.com/better-language-models/language_models_are_unsupervised_multitask_learners.pdf} {Language models are unsupervised multitask learners}.

\bibitem[{Rahman et~al.(2023)Rahman, Tabassum, Paul, Pal, and Islam}]{rahman2023bnhtrd}
Md~Ataur Rahman, Nazifa Tabassum, Manash Paul, Rajib Pal, and Mohammad~Khairul Islam. 2023.
\newblock Bn-htrd: A benchmark dataset for document level offline bangla handwritten text recognition (htr) and line segmentation.
\newblock In \emph{Computer Vision and Image Analysis for Industry 4.0}, pages 1--16. CRC Press.

\bibitem[{Redmon et~al.(2016)Redmon, Divvala, Girshick, and Farhadi}]{redmon2016yolo}
Joseph Redmon, Santosh Divvala, Ross Girshick, and Ali Farhadi. 2016.
\newblock \href {https://doi.org/10.1109/CVPR.2016.91} {You only look once: Unified, real-time object detection}.
\newblock In \emph{Proceedings of the IEEE Conference on Computer Vision and Pattern Recognition (CVPR)}, pages 779--788.

\bibitem[{Safir et~al.(2021)Safir, Ohi, Mridha, Monowar, and Hamid}]{safir2021end}
F.~B. Safir, A.~Q. Ohi, M.~F. Mridha, M.~M. Monowar, and M.~A. Hamid. 2021.
\newblock End-to-end optical character recognition for bengali handwritten words.
\newblock In \emph{2021 National Computing Colleges Conference (NCCC)}, pages 1--7.

\bibitem[{Sennrich et~al.(2016)Sennrich, Haddow, and Birch}]{sennrich2016neural}
Rico Sennrich, Barry Haddow, and Alexandra Birch. 2016.
\newblock \href {https://doi.org/10.18653/v1/P16-1162} {Neural machine translation of rare words with subword units}.
\newblock In \emph{Proceedings of the 54th Annual Meeting of the Association for Computational Linguistics (Volume 1: Long Papers)}, pages 1715--1725, Berlin, Germany. Association for Computational Linguistics.

\bibitem[{Sheng et~al.(2019)Sheng, Chen, and Xu}]{sheng2019nrtr}
F.~Sheng, Z.~Chen, and B.~Xu. 2019.
\newblock Nrtr: A no-recurrence sequence-to-sequence model for scene text recognition.
\newblock In \emph{2019 International Conference on Document Analysis and Recognition (ICDAR)}, pages 781--786. IEEE.

\bibitem[{Shi et~al.(2018)Shi, Yang, Wang, Lyu, Yao, and Bai}]{shi2018aster}
B.~Shi, M.~Yang, X.~Wang, P.~Lyu, C.~Yao, and X.~Bai. 2018.
\newblock Aster: An attentional scene text recognizer with flexible rectification.
\newblock \emph{IEEE Transactions on Pattern Analysis and Machine Intelligence}, 41(9):2035--2048.

\bibitem[{Shi et~al.(2017)Shi, Bai, and Yao}]{shi2016end2end}
Baoguang Shi, Xiang Bai, and Cong Yao. 2017.
\newblock \href {https://doi.org/10.1109/TPAMI.2016.2646371} {An end-to-end trainable neural network for image-based sequence recognition and its application to scene text recognition}.
\newblock \emph{IEEE Transactions on Pattern Analysis and Machine Intelligence}, 39(11):2298--2304.

\bibitem[{Sufian et~al.(2022)Sufian, Ghosh, Naskar, Sultana, Sil, and Rahman}]{sufian2022bdnet}
A.~Sufian, A.~Ghosh, A.~Naskar, F.~Sultana, J.~Sil, and M.~H. Rahman. 2022.
\newblock Bdnet: Bengali handwritten numeral digit recognition based on densely connected convolutional neural networks.
\newblock \emph{Journal of King Saud University - Computer and Information Sciences}, 34(6, Part A):2610--2620.

\bibitem[{Vaswani et~al.(2017)Vaswani, Shazeer, Parmar, Uszkoreit, Jones, Gomez, Kaiser, and Polosukhin}]{vaswani2017attention}
A.~Vaswani, N.~Shazeer, N.~Parmar, J.~Uszkoreit, L.~Jones, A.~N. Gomez, L.~Kaiser, and I.~Polosukhin. 2017.
\newblock Attention is all you need.
\newblock In \emph{Advances in Neural Information Processing Systems}, pages 5998--6008.

\bibitem[{Wang et~al.(2019)Wang, Wang, Qin, Zhao, and Tang}]{wang2019scene}
S.~Wang, Y.~Wang, X.~Qin, Q.~Zhao, and Z.~Tang. 2019.
\newblock Scene text recognition via gated cascade attention.
\newblock In \emph{2019 IEEE International Conference on Multimedia and Expo (ICME)}, pages 1018--1023. IEEE.

\bibitem[{Wolf et~al.(2020)Wolf, Debut, Sanh, Chaumond, Delangue, Moi, Cistac, Rault, Louf, Funtowicz, Davison, Shleifer, von Platen, Ma, Jernite, Plu, Xu, Scao, Gugger, Drame, Lhoest, and Rush}]{wolf2020huggingface}
Thomas Wolf, Lysandre Debut, Victor Sanh, Julien Chaumond, Clement Delangue, Anthony Moi, Pierric Cistac, Tim Rault, Rémi Louf, Morgan Funtowicz, Joe Davison, Sam Shleifer, Patrick von Platen, Clara Ma, Yacine Jernite, Julien Plu, Canwen Xu, Teven~Le Scao, Sylvain Gugger, and 3 others. 2020.
\newblock \href {https://doi.org/10.18653/v1/2020.emnlp-demos.6} {Transformers: State-of-the-art natural language processing}.
\newblock In \emph{Proceedings of the 2020 Conference on Empirical Methods in Natural Language Processing: System Demonstrations}, pages 38--45, Online. Association for Computational Linguistics.

\bibitem[{Zhang et~al.(2020)Zhang, Luo, Jin, Wang, Li, and Zhou}]{zhang2020sahan}
Jiaxing Zhang, Canjie Luo, Lianwen Jin, Tianwei Wang, Ziyan Li, and Weiying Zhou. 2020.
\newblock Sahan: Scale-aware hierarchical attention network for scene text recognition.
\newblock \emph{Pattern Recognition Letters}, 136:205--211.

\end{thebibliography}

\appendix

% \section{LLM Refinement}
% \label{app:refinement}

% The prompt we use to refine raw text using an LLM is provided below.
% \begin{tcolorbox}[colback=gray!8,   % light gray background
%                   colframe=black,   % black border
%                   boxrule=0.5pt,    % thin border
%                   arc=3pt,          % rounded corners
%                   left=6pt,right=6pt,top=6pt,bottom=6pt] % padding
% You are a post-processing assistant for an OCR (Optical Character Recognition) system. You will be given two inputs:
% the raw OCR output, which may contain formatting issues, spacing problems, or misrecognized characters; and an image of the original handwritten text, which is the source of truth.
% Your task is to refine the OCR output by aligning it more closely with what is actually written in the image without introducing any corrections to spelling, grammar, or wording. That means:
% Do not correct spelling mistakes or grammar, even if they appear to be incorrect.
% Do not change words based on interpretation or assumption.
% Preserve all original word choices and phrasing, even if they seem unusual.
% When in doubt, always defer to the handwritten image over the OCR output.
% Your goal is to produce an accurate, clean textual representation of the handwritten content—not a corrected or interpreted version.
% Please now refine the following OCR output using the accompanying handwritten image as reference. Output only the text found in the image, with no additional comments, explanations, or formatting.

% The OCR output follows:

% [Raw Output of our model]
% \end{tcolorbox}

\section{Synthetic Images}
\label{app:synthimages}

\begin{figure}[H]
    \centering

    \begin{subfigure}{\linewidth}
        \centering
        \includegraphics[width=0.8\linewidth]{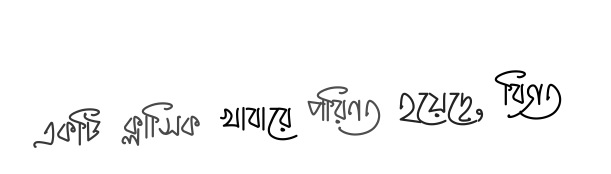}
        \caption{Bent line with a clear background.}
        \label{fig:a}
    \end{subfigure}

    \begin{subfigure}{\linewidth}
        \centering
        \includegraphics[width=0.8\linewidth]{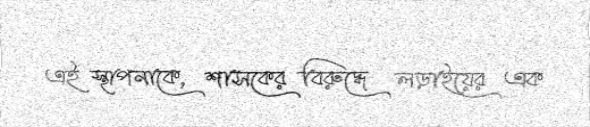}
        \caption{Line image with Gaussian blur applied.}
        \label{fig:b}
    \end{subfigure}

    \begin{subfigure}{\linewidth}
        \centering
        \includegraphics[width=0.8\linewidth]{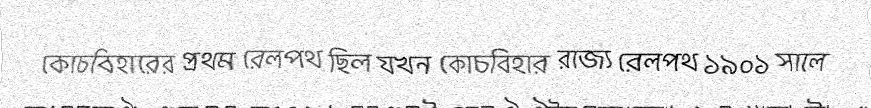}
        \caption{Wavy line with partial character fragments.}
        \label{fig:c}
    \end{subfigure}

    \begin{subfigure}{\linewidth}
        \centering
        \includegraphics[height=2cm]{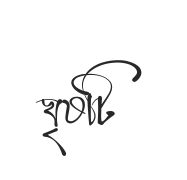}
        \caption{Word with clear background.}
        \label{fig:c}
    \end{subfigure}

    \caption{Representative examples generated by the synthetic image generator.}
    \label{fig:synthetic_images}
\end{figure}

To make synthetic images resemble real handwritten samples, we introduce artifacts such as wavy and bent text, Gaussian blur, and partial character fragments simulating real-world cropping effects during data generation. Representative examples are illustrated in Figure~\ref{fig:synthetic_images}.

% \section{Real Handwritten Finetuning Datasets}
% \label{app:real_data}
% We use five publicly available human-labeled datasets to fine-tune our word-level models: BN-HTRd~\citep{rahman2023bnhtrd}, BanglaWriting~\citep{mridha2021banglawriting}, BanglaLekha-Isolated~\citep{biswas2017banglalekha}, IIIT-Indic~\citep{jindal2021iiitindic}, and ICDAR 2024~\citep{icdar2024dataset}. Details of these datasets are provided in Table~\ref{tab:dataset_statistics}.

% \begin{table}[H]
%     \centering
%     \small
%     \begin{tabular}{l r}
%         \toprule
%         \textbf{Dataset} & \textbf{\# Words} \\
%         \midrule
%         BN-HTRd                & 104,854 \\
%         BanglaWriting          & 21,234 \\
%         BanglaLekha-Isolated (Digits only)  & 19,748 \\
%         IIIT-Indic             & 113,075 \\
%         ICDAR 2024             & 79,663 \\
%         \midrule
%         \textbf{Total} & \textbf{338,574} \\
%         \bottomrule
%     \end{tabular}
%     \caption{Summary of human-labeled datasets used to fine-tune the word-level models.}
%     \label{tab:dataset_statistics}
% \end{table}

\section{Definitions of the Evaluation Metrics}
\label{app:evalmetrics}
We provide the definitions of Character Error Rate (CER) and Word Error Rate (WER) below.

\paragraph{CER.}
CER measures the proportion of character-level errors—substitutions ($S$), deletions ($D$), and insertions ($I$)—required to transform the predicted text sequence (hypothesis) into the reference text (ground truth), normalized by the total number of characters in the reference ($N$). 
Formally,
$$
\mathrm{CER} = \frac{S + D + I}{N}
$$

\paragraph{WER.}
WER measures the proportion of word-level errors—substitutions ($S$), deletions ($D$), and insertions ($I$)—needed to convert the predicted word sequence into the reference word sequence, divided by the total number of words in the reference ($N$). 
Formally,
$$
\mathrm{WER} = \frac{S + D + I}{N}
$$

\section{Training Details}
\label{app:traindetails}
% Our model was pretrained and fine-tuned on a single NVIDIA GeForce RTX 3080 GPU with a batch size of 32. The first-stage pretraining on 4.5M synthetic line-level images required about 6 hours per epoch, while the second-stage pretraining on 7M synthetic word-level images required about 9 hours per epoch. We pretrained for 2 epochs and 1 epoch in the first and second stage, respectively, followed by finetuning for 4 epochs on real handwritten datasets.

Our model was pretrained and fine-tuned on a single NVIDIA GeForce RTX 3080 GPU with a batch size of 32. The first-stage pretraining on 4.5M synthetic line-level images requires 13 hours over 2 epochs, while the second-stage pretraining on 7M synthetic word-level images takes 9 hours for a single epoch. The model is then fine-tuned on real handwritten datasets for 4 epochs, completing in less than 2 hours.

\section{LLM Evaluation}
\label{app:llmeval}

The prompt we use to transcribe an image using an external LLM is provided below.
\begin{tcolorbox}[colback=gray!8,   % light gray background
                  colframe=black,   % black border
                  boxrule=0.5pt,    % thin border
                  arc=3pt,          % rounded corners
                  left=6pt,right=6pt,top=6pt,bottom=6pt] % padding
You are an optical character recognition (OCR) tool. Your task is to transcribe the handwritten Bengali text shown in the provided image directly and exactly, without making any corrections, translations, or modifications. Extract the text precisely as it appears, preserving original spelling, grammar, and formatting.

Output only the text found in the image, with no additional comments, explanations, or formatting.

\end{tcolorbox}

\section{Illustrative Outputs from the System}
\label{app:sampleoutputs}

\begin{figure*}[!t]
    \centering

    \begin{subfigure}[t]{\textwidth}
        \centering
        \includegraphics[width=\textwidth]{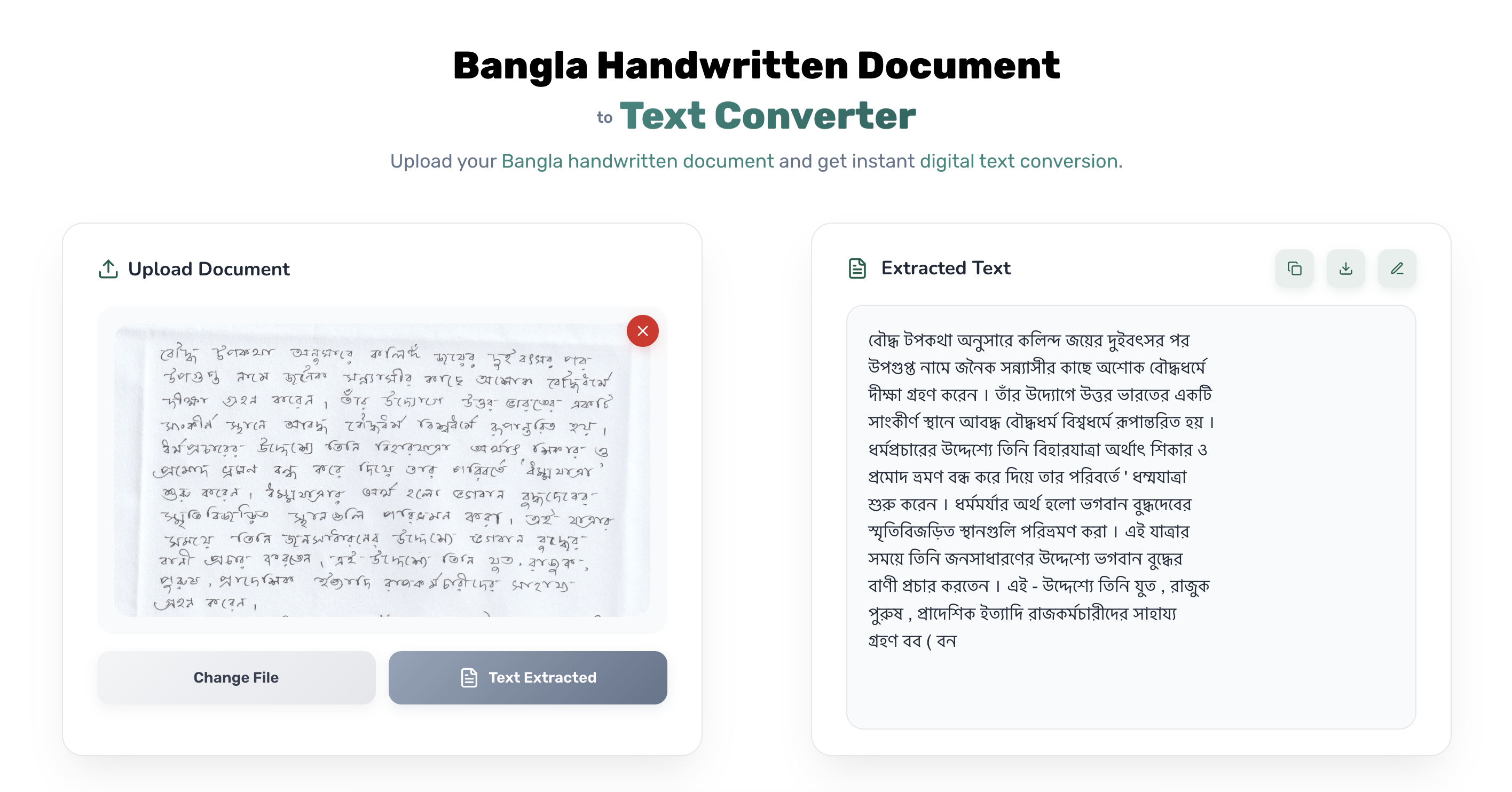}
        \caption{Example of the interface after uploading an image.}
        \label{fig:interface_image_upload}
    \end{subfigure}
    
    \vspace{1em}

    \begin{subfigure}[t]{\textwidth}
        \centering
        \includegraphics[width=\textwidth]{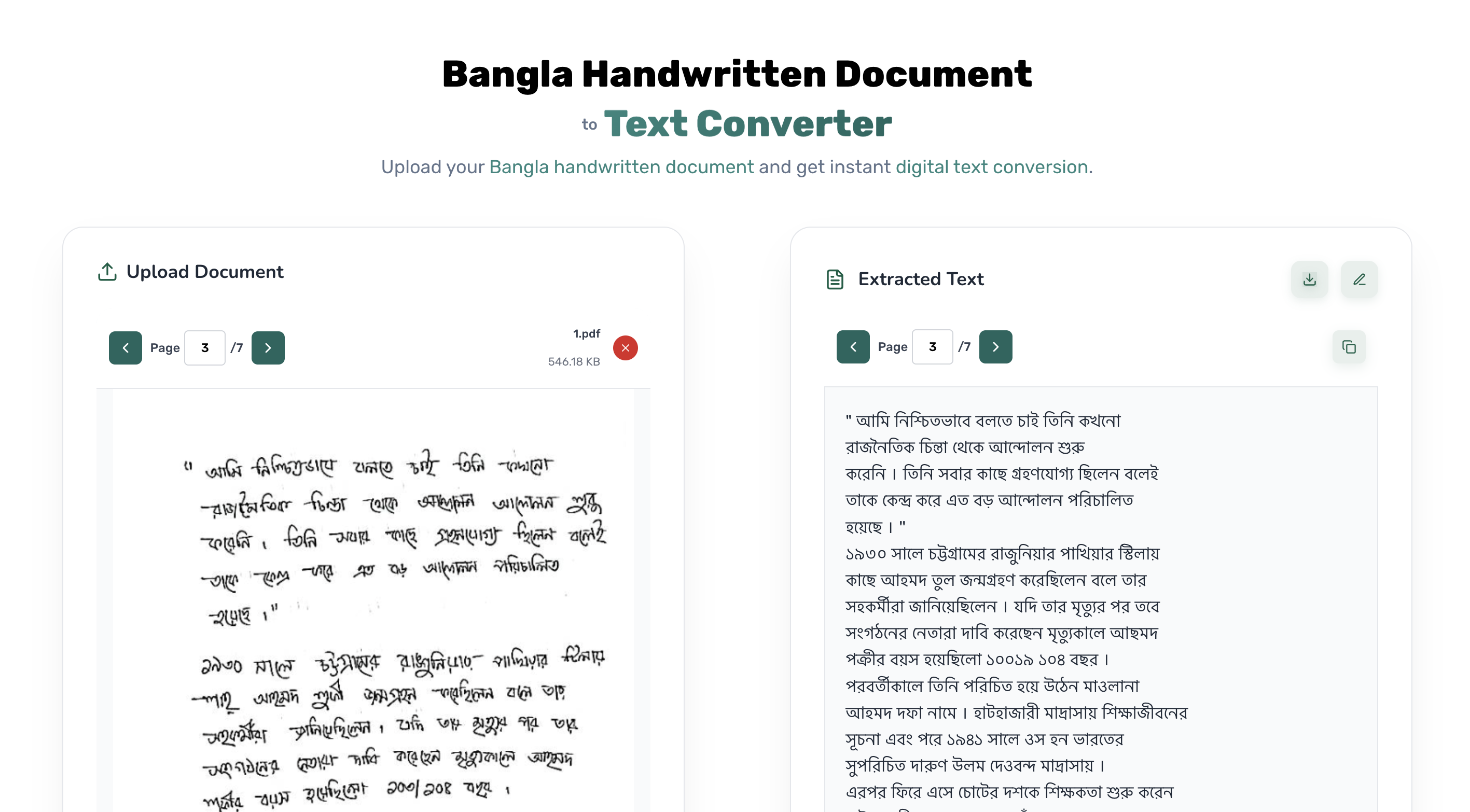}
        \vspace{1em} % Add space between second image and its caption
        \caption{Example of the interface after uploading a PDF.}
        \label{fig:interface_pdf_upload}
    \end{subfigure}

    \caption{Screenshot of the Bangla Handwritten Document to Text Converter web interface, showing the document upload area on the left and the extracted text display area on the right.}
    \label{fig:interface}
\end{figure*}

Figure~\ref{fig:interface} depicts the system’s user interface along with sample use-cases. A video demonstrating how to use the system is available on YouTube\footnote{\url{https://youtu.be/ckgWBHQarxc}}.

\end{document}